\documentclass[twoside]{article}

\usepackage[accepted]{aistats2024}
\usepackage[round]{natbib}

\usepackage{bm}
\usepackage{mathtools}
\usepackage{amsthm}
\usepackage{amsmath}
\usepackage{amsfonts}
\DeclareMathOperator*{\argmin}{arg\,min}
\usepackage[capitalize,noabbrev]{cleveref}
\RequirePackage{algorithm}
\RequirePackage{algorithmic}
\usepackage{booktabs}

\theoremstyle{plain}

\newtheorem{proposition}{Proposition}
\newtheorem{lemma}{Lemma}

\theoremstyle{definition}
\newtheorem{definition}{Definition}

\theoremstyle{remark}

\begin{document}

%

%

\twocolumn[

\aistatstitle{Scalable Higher-Order Tensor Product Spline Models}

\aistatsauthor{ David R\"ugamer }

\aistatsaddress{ Departments of Statistics, LMU Munich\\ Munich Center for Machine Learning } ]

\begin{abstract}
In the current era of vast data and transparent machine learning, it is essential for techniques to operate at a large scale while providing a clear mathematical comprehension of the internal workings of the method. Although there already exist interpretable semi-parametric regression methods for large-scale applications that take into account non-linearity in the data, the complexity of the models is still often limited. One of the main challenges is the absence of interactions in these models, which are left out for the sake of better interpretability but also due to impractical computational costs. To overcome this limitation, we propose a new approach using a factorization method to derive a highly scalable higher-order tensor product spline model. Our method allows for the incorporation of all (higher-order) interactions of non-linear feature effects while having computational costs proportional to a model without interactions. We further develop a meaningful penalization scheme and examine the induced optimization problem. We conclude by evaluating the predictive and estimation performance of our method.
\end{abstract}

\section{INTRODUCTION}

Two of the core principles of statistical regression models are additivity and linearity of the predictors. These properties allow estimated feature effects to be easily interpreted, which also led to (revived) interest in such models in the machine learning and information retrieval community \citep{chen2017group, Wang.2020, Zhuang.2021, chang2021nodegam}. A frequently used and cited example of an interpretable yet flexible statistical regression model is the generalized additive model \citep[GAM;][]{hastie2017generalized, Wood.2017.book}. Using basis functions 
to approximate non-linear functions, 
these models can represent non-linear feature effects in one or a moderate number of dimensions. Applying this principle in settings with many features and higher-order interactions, however, comes with considerable downsides. For univariate non-linear effects, the number of basis functions $M$ for each feature typically lies in the range of 10 to 20 and needs to be evaluated prior to model fitting. Representing and fitting all available features using basis functions will not only result in a notable increase in training time, but also requires a considerable amount of additional memory. In higher dimensions $D$, these problems carry even more weight as $D$-variate non-linear representations are typically constructed using Kronecker or tensor product splines (TPS), i.e., a (row-wise) Kronecker product of all involved bases. This results in computational costs of $\mathcal{O}((pM)^D)$ for TPS models with $p$ features. 
\begin{figure}[!t]
\begin{center}
\centerline{\includegraphics[width=\columnwidth]{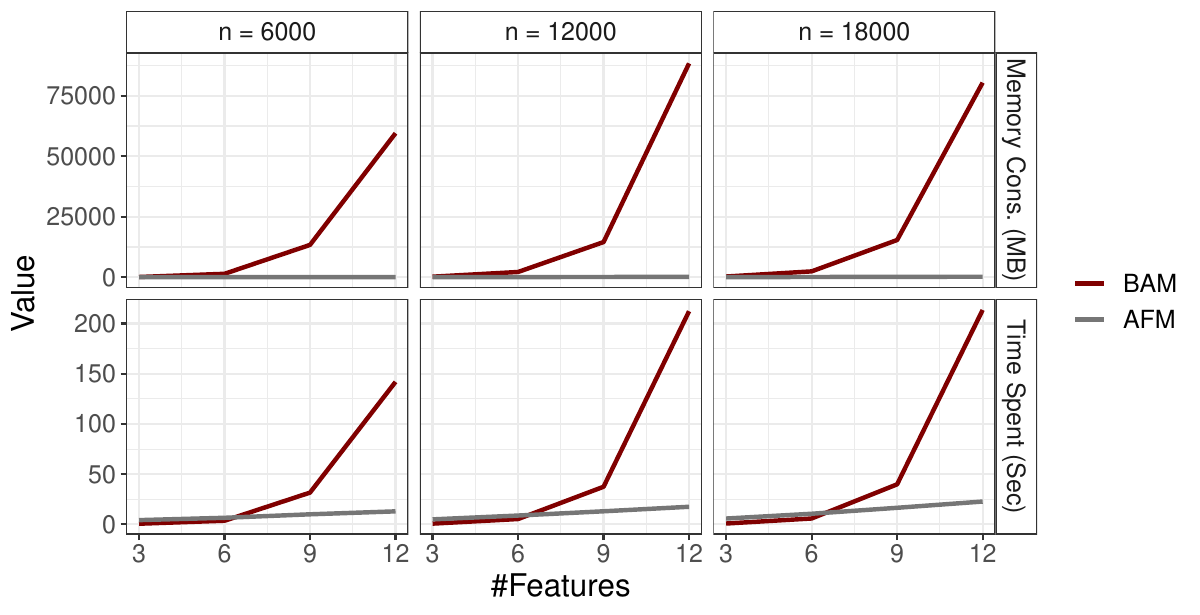}}
\vskip -0.15in
    \caption{\small Comparison of memory consumption (first row) and time consumption (second row) between the state-of-the-art big additive model (BAM) implementation (in red) and our proposal (in gray) when fitting a model for all $\binom{p}{2}$ tensor product splines using different numbers of features $p$ (x-axis) and observations (columns).}
    \label{fig:mem}
\end{center}
\vskip -0.3in
\end{figure}
Computational feasibility is thus one of the main reasons statistical applications are often restricted to only uni- and bivariate (tensor product) splines. While several approaches to tackle this problem have been proposed \citep[e.g.,][]{Wood.2017}, existing solutions still either suffer from extensive memory or runtime costs.


\textbf{Our Contribution}: In order to efficiently scale additive models in higher dimensions (cf. also Figure~\ref{fig:mem}), we propose an approach for modeling higher-order TPS that only has linear complexity in the number of features $p$ based on the idea of factorization machines \citep[FMs;][]{Rendle.2010}.  While this effectively addresses existing scaling problems of GAMs, our approach also extends (higher-order) factorization machines by allowing for non-linear relationships. In addition to deriving the resulting computational complexity, we also propose an efficient way of computing the model, suggest a suitable penalization scheme and provide an optimization routine for our approach. Our experimental section demonstrates that models with higher-order TPS work well in practice, show desirable approximation behavior, and yield competitive results on real-world data. 

\section{RELATED LITERATURE}

\paragraph{GAMs and TPS} GAMs extend generalized linear models \citep{Nelder.1972} by allowing feature effects to be non-linear, typically achieved by using a spline bases representation. Next to basic principles \citep[see, e.g.,][]{hastie2017generalized, Wood.2017.book}, many extensions of GAMs have been discussed in the literature. In order to include non-linear functions of more than one variable, various options exist, e.g., by using spline bases in multiple dimensions. \cite{Wood.2006} proposed a flexible way of constructing multivariate non-linear functions in GAMs using TPS as an alternative option, which we will outline in more detail in Section~\ref{sec:gams_tp}. Although GAM software is usually optimized in terms of efficiency, computational costs can be a bottleneck for large data sets or complex model formulations.  While there exist approaches work that allows GAM estimation for data with many observations \citep{Wood.2017}, GAMs still scale unfavorably with many features or feature interactions. Recent approaches \citep[e.g.,][]{Ruegamer.2020,pho} suggest fitting structured regression models as (part of) a neural network. This can result in a better space complexity in situations with many data points and allows for more flexibility in the additive predictors of models. 

\paragraph{Factorization Approaches} Similar to GAMs, factorization approaches have been studied extensively. Popularized for recommender systems, different (matrix) factorization approaches have been proposed in the early 2000s \citep[see, e.g.,][]{Srebro.2004, Adomavicius.2005, Koren.2009} and are still considered state-of-the-art in terms of performance and efficiency \citep{Rendle.2020, Jin.2021}. Factorization approaches are also used as part of additive models in order to improve scalability \citep{fdtf2021}. Closely related to matrix factorization are factorization machines \citep[FMs;][]{Rendle.2010}. FMs are based on a linear model formulation with pairwise interactions between all features and use a factorization trick to overcome unfavorable scaling when the number of features is large. Various extensions have been developed over the past years, including convex FMs \citep{blondel2015convex} and an efficient calculation of higher-order FMs \citep[HOFMs;][]{blondel2016higher}. 
Other extensions include boosted FMs \citep{Yuan.2017}, FMs with special personalized feature treatment \citep{Chen.2019} or interaction-aware FMs \citep{Hong.2019}. Similar to our proposal, \cite{lan2019accurate} use a non-parametric subspace feature mapping to encode interactions and account for non-linearity, but rely on binning the features. 

\paragraph{Boosting} Apart from FMs and GAMs, various other approaches exist that model non-linearity and/or interactions while preserving an additivity structure of the model. One of the most prominent approaches in machine learning that combines additivity and predictive performance is boosting. Already in the seminal work of Friedman \citep{friedman2001greedy}, (functional gradient) boosting was proposed to optimize additive models (potentially with interactions). This idea is used, e.g., for GAMboost \citep{mboost} and lays the foundation for other interpretable boosting frameworks such as GA${}^2$M \citep{lou2013accurate}, explainable boosting machines \citep{nori2019interpretml}, or functional regression boosting \citep{brockhaus2017boosting}. While extensions for scalability have been suggested \citep[see, e.g.][]{Schalk.2021}, boosting GAMs with all possible interaction terms remains challenging. 


\section{BACKGROUND} \label{sec:background}

We first introduce our notation and then give a short introduction into GAMs in Section~\ref{sec:gams}. For more details, see, e.g., \cite{Wood.2017.book}.


In the following, we write scalar values in small or capital letters without formatting, vectors in small bold letters, matrices in capital bold letters, and tensors using fraktur typeface, e.g., $\mathfrak{X}$. Calligraphic letters will have different meaning depending on the context, while $\mathcal{O}$ is reserved to describe the complexity of calculations in terms of computing time or memory. 
For better readability, we will denote the sequence from $1,\ldots,x$ with $[x]$. We further use $\otimes$ for the Kronecker product. 



\subsection{Generalized Additive Models} \label{sec:gams}

Given the response random variable $Y$ and $p$ features $\bm{x} = (x_1,\ldots,x_p)$, an additive model with linear and non-linear effects for all features assumes the following relationship:
\begin{equation} \label{eq:unigam}
    Y = \eta(\bm{x}) + \varepsilon = \alpha_0 + \sum_{j=1}^p x_j \alpha_j + \sum_{j=1}^p f_j(x_j) + \varepsilon,
\end{equation}
where $\alpha_0, \alpha_1, \ldots, \alpha_p$ are linear regression coefficients, $f_1,\ldots,f_p$ univariate non-linear functions, $\varepsilon\sim \mathcal{N}(0,\sigma^2)$ is a zero-mean Gaussian random variable with variance $\sigma^2 > 0$ and $\eta$ the model predictor. GAMs, the generalization of additive models, replace the distribution assumption in \eqref{eq:unigam} using a more general distribution by assuming that $Y|\bm{x}$ has some exponential family distribution and $\mathbb{E}(Y|\bm{x}) = h(\eta(\bm{x}))$ for some monotonic (response) function $h$. A prediction $\hat{y} = h(\hat{\eta}(\bm{x}))$ for the observed value $y$ in GAMs is formed by estimating the regression coefficients and functions $f_j$. GAMs can be optimized using (different types of) maximum likelihood estimation. Alternatively, using the negative log-likelihood as (convex) loss function $\ell$, their optimization can also be framed as an empirical risk minimization problem.

Since linear effects $\alpha_j$ can be incorporated in the functions $f_j$, we will drop the linear model part in the following. The non-linear functions $f_j$ in GAMs are usually approximated using a (spline) basis representation, i.e., 
\begin{equation} \label{eq:basisapprox}
f_j(x_j) \approx \sum_{m=1}^{M_j} B_{m,j}(x_j) \beta_{m,j} = \bm{B}_j^\top \bm{\beta}_j,    
\end{equation}
where $B_{m,j}$ are pre-defined basis functions (e.g., truncated polynomials or B-splines) and $\beta_{m,j}$ the corresponding basis coefficients. We summarize all basis functions and coefficients using $\bm{B}_j = (B_{1,j},\ldots,B_{M_j,j}) \in \mathbb{R}^{M_j}$ and $\bm{\beta}_j = (\beta_{1,j},\ldots,\beta_{M_j,j})^\top \in \mathbb{R}^{M_j}$, respectively. To enforce smoothness of the functions $f_j$, the $\bm{\beta}_j$ coefficients are typically estimated using a smoothness penalty.

\subsubsection{Smoothness Penalties} \label{sec:gams_pen}

One of the most common approaches to estimate smooth functions $f_j$ is to employ a difference penalty for successive basis coefficients $\beta_{m,j}, \beta_{m+1,j}$ of basis functions $B_{m,j}(x)$, $B_{m+1,j}(x)$, which penalizes deviating behavior in neighboring basis functions. The penalty term for the penalized loss function is then given by 
     $\mathcal{P} = \sum_{j=1}^p \lambda_j \int (f_j''(x))^2\, \mathrm{d}x$,
which is a trade-off between goodness-of-fit and roughness of the functions $f_j$. The penalized loss can be written as
\begin{equation}
    \ell(y,\hat{y}) + \sum_{j=1}^p \lambda_j \bm{\beta}_j^\top \bm{P}_j \bm{\beta}_j,
\end{equation}
where $\bm{P}_j \in \mathbb{R}^{M_j \times M_j}$ is a squared penalty matrix depending on the evaluated basis $\bm{B}_j$ for the $j$th feature and usually penalizes first or second differences in the coefficients $\bm{\beta}_j$.

GAMs also allow for higher dimensional non-linear functions, e.g., bivariate smooth terms $f_{k,l}(x_k,x_l)$. A common approach for their construction are tensor product splines.

\subsubsection{Tensor Product Splines} \label{sec:gams_tp}

While there are various approaches to model smooth functions of several features, TPS constructed from marginal univariate bases constitute an attractive option. The resulting smooth terms are very flexible, scale-invariant, relatively low rank as well as easy to construct and interpret \citep[see][]{Wood.2006}. For a model with all $\binom{p}{2}$ possible bivariate effects, the TPS part is given by
\begin{equation}
\begin{split} \label{eq:bigam}
    \sum_{k=1}^p &\sum_{l=k+1}^{p} f_{k,l}(x_k, x_l) \approx \sum_{k=1}^p \sum_{l=k+1}^{p} (\bm{B}_k \otimes \bm{B}_l) \bm{\beta}_{k,l}\\ 
    &= \sum_{k=1}^p \sum_{l=k+1}^{p} \sum_{m=1}^{M_k} \sum_{o=1}^{O_l} B_{m,k}(x_k) B_{o,l}(x_l) \beta_{m,k,o,l}
\end{split}    
\end{equation}
with univariate spline basis functions $B_{m,k}$, $B_{o,l}$ and basis coefficients $\beta_{m,k,o,l}$, summarized in $\bm{\beta}_{k,l} \in \mathbb{R}^{M_k O_l}$. Bivariate TPS are penalized using
\begin{equation*}
    \mathcal{P}(f_{k,l}) = \int_{x_k, x_l} \lambda_{k} (\partial^2 f / \partial x_k^2)^2 + \lambda_{l} (\partial^2 f / \partial x_l^2)^2 \, \mathrm{d}x_k x_l,
\end{equation*}
which can be written as 
\begin{equation} \label{eq:bivpen}
    \mathcal{P}(f_{k,l}) = \bm{\beta}_{k,l}^\top (\lambda_{k} \bm{P}_k \oplus \lambda_l \bm{P}_l) \bm{\beta}_{k,l},
\end{equation}
where $\oplus$ is the Kronecker sum operator.

This principle can be generalized to $D$-variate smooths for variables $\mathcal{J} := \{j_1, \ldots, j_D\}$, which are approximated by 

\begin{equation} \label{eq:dvarspline}
f_{j_1,\ldots,j_D}(x_{j_1},\ldots,x_{j_D}) \approx (\otimes_{j\in\mathcal{J}} \bm{B}_j) \bm{\beta}_{\mathcal{J}},
\end{equation}
where $\bm{\beta}_{\mathcal{J}} \in \mathbb{R}^{\prod_{t=1}^D M_{j_t}}$ contains the coefficients for all combinations of the $D$ basis functions. The corresponding penalty term for \eqref{eq:dvarspline} is constructed analogously to \eqref{eq:bivpen} by $\bm{\beta}_{\mathcal{J}}^\top (\oplus_{j\in\mathcal{J}} \lambda_j \bm{P}_j) \bm{\beta}_{\mathcal{J}}$.

In the following, we assume that the number of spline basis functions is roughly equal across different features and use $M := \max_{k,l}\{M_k,O_l\}$ to denote the spline basis in multivariate splines that uses the most basis functions.

\section{SCALABLE HIGHER-ORDER TENSOR PRODUCT SPLINE MODELS} \label{sec:proposal}

As can be directly inferred from \eqref{eq:bigam}, the cost of fitting a bivariate TPS is $\mathcal{O}(p^2  M^2)$. While $M$ is usually kept fixed and of moderate size (e.g., $M = 10$), this implies that models will be increasingly expensive for both a growing number of basis evaluations and number of features $p$. For models with (up to) $D$-variate TPS, the computational cost increases to $\mathcal{O}(p^D M^D)$. This makes GAMs infeasible both in terms of computing time and also in terms of memory storage.

\subsection{Additive Factorization Machines}

To overcome the unfavorable scaling of GAMs with many (or higher-order) TPS, we introduce additive factorization machines (AFMs). Based on the idea of factorization machines, we approximate $f_{k,l}(x_k, x_l)$ in \eqref{eq:bigam} by $\phi_{k,l}(x_k, x_l)$ defined as
\begin{equation} \label{eq:afmapprox}
\begin{split}
    \phi_{k,l}&(x_k, x_l) = \\
&\sum_{m=1}^{M_k} \sum_{o=1}^{O_l} B_{m,k}(x_k) B_{o,l}(x_l) \sum_{f=1}^F \gamma_{m,k,f} \gamma_{o,l,f},
\end{split}
\end{equation}
where ${\gamma}_{\cdot,k,f} \in \mathbb{R}$ are latent factors approximating the joint effect $\beta_{m,k,o,l}$. When approximating every bivariate interaction term in \eqref{eq:bigam} with the term defined in \eqref{eq:afmapprox}, we can derive the following representation.
\begin{lemma}[AFM Representation] \label{th:afmrep}
The approximation of \eqref{eq:afmapprox} using \eqref{eq:bigam} can be written as
\begin{equation} \label{eq:biafmapprox}
\begin{split} 
    \sum_{k=1}^p  \sum_{l=k+1}^{p}\!\! f_{k,l}(x_k, x_l) \approx \frac{1}{2} \sum_{f=1}^F \left\lbrace \!\left[ \sum_{k=1}^p \varphi_{k,f} \right]^2\!\! - \sum_{k=1}^p \varphi_{k,f}^2  \right\rbrace\!,
\end{split}    
\end{equation}
with $\varphi_{k,f} = \sum_{m=1}^{M_k} B_{m,k}(x_k) \gamma_{m,k,f}$.
\end{lemma}

As a direct result of Lemma~\ref{th:afmrep}, we obtain the scaling of computing AFMs.
\begin{proposition}[Linear Scaling of AFMs]\label{th:afmscale}
Computations for AFMs scale with $\mathcal{O}(p  M  F)$.
\end{proposition}
Lemma~\ref{th:afmrep} and Proposition~\ref{th:afmscale} are natural extensions of linearity results from FMs. A proof of the corollary is provided in the Supplementary Material. Roughly speaking, the factorization trick from FMs also works for AFMs in a similar manner as the additional basis function dimension only depends on the respective feature dimension. In particular, this means that AFMs scale linearly both in the number of features $p$ and the spline basis dimension $M$. Another direct result of this representation and noteworthy property unique to AFMs is given in the following proposition for a data set of $n$ observations.
\begin{proposition}[Basis Evaluations in AFMs] \label{th:afmstore}
If every feature in AFMs is represented by only one basis, it suffices to evaluate all univariate basis functions once for each feature and the memory costs for storing all features are $\mathcal{O}(n p  M)$.
\end{proposition}
While this seems inconspicuous at first glance, a na\"{i}ve approach for bivariate models requires storing $\mathcal{O}(n  p^2  M^2)$ entries, which is infeasible if $n$ or $p$ is large. In contrast, AFMs only require the same amount of storage as for a univariate spline model. Note that the number of parameters also (linearly) increases with $F$ and needs to be taken into account for the total required storage. However, Proposition~\ref{th:afmstore} specifically looks at the costs of storing basis evaluated features in memory as this can be a storage bottleneck during the pre-processing of GAMs. 

\subsection{Additive Higher-Order Factorization Machines} \label{sec:hofm}

We now extend previous results to the general case of a $D$-variate interaction GAM. By analogy to higher-order FMs \citep{blondel2015convex}, we refer to our approximation as \emph{additive higher-order factorization machines} (AHOFMs). The goal of AHOFMs is to provide a scalable version of a TPS model which includes multivariate splines up to $D$-variate smooths, i.e.,
\begin{equation} \label{eq:dvargam}
\begin{split}
    \eta(\bm{x}) = &\alpha_0 + \sum_{j=1}^p f_j(x_j) + \sum_{j' > j} f_{j',j}(x_{j'}, x_j) + \ldots\\ 
    &+ \sum_{j_D > \cdots > j_1} f_{j_1,\ldots,j_D}(x_{j_1},\ldots,x_{j_D}).
\end{split}
\end{equation}
Assume a TPS representation for all smooth terms in \eqref{eq:dvargam} with a maximum number of $M$ basis functions. Then, the cost of computing only the last term in \eqref{eq:dvargam} is already $\mathcal{O}(p^D  M^D)$ (and analogous for memory costs).  

Inspired by Vieta's formula and the ANOVA kernel, we can derive a similar result as given in \cite{blondel2016higher} to reduce the cost of computing the $d$th degree term in AHOFMs. We will make the degree $d$ explicit for $\gamma$ and $\varphi$ using the superscript ${(d)}$.
\begin{definition}[Additive Higher-order Term (AHOT)]
The $f$th additive higher-order term (AHOT) of degree $2 \leq d \leq D$ in AHOFMs is given by
\begin{equation} \label{eq:AHOT}
\Phi^{(d)}_f = \sum_{j_d > \cdots > j_1} \prod_{t=1}^d \sum_{m=1}^{M_{j_t}} B_{m,j_t}(x_{j_t}) \gamma^{(d)}_{m,j_t,f}.
\end{equation}
\end{definition}
We use $F_d$ AHOTs to approximate a $d$-variate smooth:
\begin{equation}
\sum_{f=1}^{F_d} \Phi^{(d)}_f \approx \sum_{j_d > \cdots > j_1} f_{j_1,\ldots,j_d}(x_{j_1},\ldots,x_{j_d}).
\end{equation}
and estimate the $D$-variate TPS model \eqref{eq:dvargam} with an AHOFM of degree $D$, defined as follows.
\begin{definition}[AHOFM of Degree $D$] \label{eq:AHOFMs}
The predictor $\eta(\bm{x})$ of an AHOFM of degree $D$ is defined by
\begin{equation}
        \alpha_0 + \sum_{j=1}^p B_{m,j}(x_j) \beta_{m,j} + \sum_{d=2}^D \sum_{f=1}^{F_d} \Phi^{(d)}_f.
\end{equation}
\end{definition}
The following corollary defines how to recursively describe all AHOTs for $d\geq 2$ using univariate spline representations $\varphi_{j,f}$ as defined in \eqref{eq:biafmapprox}. 
\begin{lemma}[Representation AHOT of Degree $d$] \label{th:ahot}
Let $\Phi^{(0)}_f \equiv 1$ as well as $\Phi^{(1)}_f = \sum_{j=1}^p \varphi^{(d)}_{j,f}$. The degree $d\geq 2$ AHOT can be recursively defined by
\begin{equation} \label{eq:recursion}
    \Phi_f^{(d)} = \frac{1}{d} \sum_{t=1}^d (-1)^{t+1} \Phi_f^{(d-t)} \left\lbrace \sum_{j=1}^p \left[\varphi^{(d)}_{j,f}\right]^t \right\rbrace.
\end{equation}
\end{lemma}

The recursive representation \eqref{eq:recursion} allows us to efficiently calculate AHOTs of higher order. The corresponding proof can be found in the Supplementary Material. As another consequence of this representation, we have the following scaling properties.
\begin{proposition}[Linear Scaling of AHOFMs] \label{th:ahofmscale}
Computations for AHOFMs scale with $\mathcal{O}(p  M  \mathcal{F}  D + \mathcal{F}  D^2)$ where $\mathcal{F} = \sum_{d=1}^D F_d$.
\end{proposition}
Since $D$ is usually small, computations again roughly scale linearly with the number of features, the basis, and the latent factor dimension. We also recognize that despite the increased dimension $D$, every feature basis has to be evaluated only once.
\begin{proposition}[Basis Evaluations in AHOFMs] \label{th:ahofmstore}
If every feature in AHOFMs is represented by only one basis, it suffices to evaluate all univariate basis functions once for each feature and the memory costs for storing all features are $\mathcal{O}(n p  M)$.
\end{proposition}
While the memory consumption also increases with $\mathcal{F}$ when considering the storage of all $\gamma$ parameters, Proposition~\ref{th:ahofmstore} again focuses on the storage of all features after applying the basis evaluations. Similar to the kernel trick, higher-order features are not actually calculated and stored in memory, as AHOFMs only work on the (basis evaluated) univariate features independent of $D$.

In contrast to FMs and HOFMs, AHOFMs require additional considerations to enforce appropriate smoothness of all non-linear functions in \eqref{eq:dvargam} without impairing favorable scaling properties. 

\subsection{Penalization and Optimization} \label{sec:pen}

Learning the smoothness of many (higher-order) TPS is an additional challenge discussed below. 

\subsubsection{Penalization}

Following the penalization scheme of TPS described in \cref{sec:gams_tp}, we propose a smoothness penalization for AHOFMs based on the penalties of involved marginal bases $B_{m,j}$ with corresponding difference penalty matrices $\bm{P}_j$. Let $\mathfrak{G}^{(d)}$ be the array of all coefficients $\gamma^{(d)}_{m,j,f}$ for all $m\in[M],j\in[p],f\in[F_d]$. For simplicity\footnote{Next to a ragged tensor definition that allows for a varying first dimension, padding the tensor can also be an option to always have $M$ dimensions for every feature $j$.}, we assume $M_j \equiv M$, so that $\mathfrak{G}^{(d)} \in \mathbb{R}^{M \times p \times F_d}$. Further, let $\mathfrak{G}_{[D]} = \mathfrak{G}^{(1)}, \ldots, \mathfrak{G}^{(D)}$ and $\bm{\gamma}^{(d)}_{j,f} = (\gamma^{(d)}_{1,j,f},\ldots,\gamma^{(d)}_{M,j,f})^\top$, i.e., the Mode-1 (column) fibers of $\mathfrak{G}^{(d)}$, and $\bm{\Theta} = (\alpha_0, \bm{\beta}_1, \ldots, \bm{\beta}_p)$. 

We define the penalty of AHOFMs as follows.
\begin{definition}[AHOFM Penalty] \label{def:ahofmpen}
The smoothing penalty of AHOFMs is defined as
\begin{equation} 
\begin{split}
    \mathcal{P}(\mathfrak{G}_{[D]},\bm{\Theta}) = &\sum_{j=1}^p \lambda_j \bm{\beta}_j^\top \bm{P}_j \bm{\beta}_j +\\ 
    &\sum_{d=2}^D \sum_{f=1}^{F_d} \sum_{j=1}^p \lambda^{(d)}_{j,f} {\bm{\gamma}^{(d)}_{j,f}}^\top \bm{P}_j \bm{\gamma}^{(d)}_{j,f}.
    \end{split}
\end{equation}
\end{definition}
Due to the independence assumption of all latent factors involved in every factorization, it is natural to only penalize univariate directions as expressed in \cref{def:ahofmpen}. A regularization that involves multiple dimensions would further result in a non-decomposable penalty w.r.t.~the $\bm{\gamma}^{(d)}_{j,f}$ and make the optimization of AHOFMs more challenging (see Section~\ref{sec:optim} for details). The penalized optimization problem for $n$ i.i.d.~data points $(y_i, \bm{x}_i)_{i\in[n]}$ with $\bm{x}_i = (x_{i,1},\ldots,x_{i,p})$ and loss function $\ell$ is then given by \begin{equation} \label{eq:penloss}
    \argmin_{\mathfrak{G}_{[D]}, \bm{\Theta}} \sum_{i=1}^n \ell(y_i, \eta_i(\bm{x})) + \frac{1}{2} \mathcal{P}(\mathfrak{G}_{[D]}, \bm{\Theta}).
\end{equation}
In \eqref{eq:penloss}, the smoothing parameters are considered to be tuning parameters. While it is possible in univariate GAMs to estimate the smoothing parameters $\lambda_1,\ldots,\lambda_p$ directly, this becomes computational challenging for higher-order TPS due to the exponentially increasing amount of parameters. HOFMs avoid this combinatorial explosion of hyperparameters by setting all the parameters to the same value \citep{blondel2016higher}. This, however, is not a meaningful approach for smoothing parameters as every smooth term can potentially live on a completely different domain (e.g., $\lambda = 1$ can imply no penalization for one smooth, but maximum penalization for another term).

\subsubsection{Scalable Smoothing}

To derive a meaningful penalization in AHOFMs, we exploit the definition of \emph{degrees-of-freedom} for penalized linear smoothers \citep{Buja.1989}. Given a matrix of basis evaluations $\bm{B}_j\in\mathbb{R}^{n\times M_j}$ with entries $B_{m,j}(x_{i,j})$ and a squared penalty matrix $\bm{P}_j$ for the penalization of differences in neighboring basis coefficients, there exists a one-to-one map between $\lambda_{j,f}^{(d)}$ and the respective degrees-of-freedom 
\begin{equation} \label{eq:df}
\text{df}^{(d)}_{j,f}(\lambda_{j,f}^{(d)}) = \text{tr}(2\bm{H}_j(\lambda_{j,f}^{(d)}) - \bm{H}_j(\lambda_{j,f}^{(d)})^\top\bm{H}_j(\lambda_{j,f}^{(d)}))
\end{equation}
with $\bm{H}_j(\lambda_{j,f}^{(d)}) = \bm{B}_j (\bm{B}_j^\top \bm{B}_j + \lambda_{j,f}^{(d)} \bm{P}_j)^{-1} \bm{B}_j^\top$. While the exact degrees-of-freedom only hold for a linear model with a single smooth term, this approach allows to define a meaningful a priori amount of penalization for all smooth terms by restricting their degrees-of-freedom to the same global $\text{df}^{(d)}$ value as follows.
\begin{proposition}[Homogeneous AHOFM Smoothing] \label{th:homo}
Given a global $\text{df}^{(d)}$ value, an equal amount of penalization for all $\binom{p}{d}$ AHOTs in $\Phi_{f}^{(d)}$ is achieved by choosing $\lambda^{(d)}_{j,f}$ such that $\text{df}_{j,f}^{(d)}(\lambda_{j,f}^{(d)}) \equiv \text{df}^{(d)}\forall j\in[p],f\in[F_d]$.
\end{proposition}
The Demmler-Reinsch orthogonalization \citep{ruppert2003semiparametric} can be used to efficiently solve \eqref{eq:df} for $\lambda_{j,f}^{(d)}$, i.e., calculate $\lambda_{j,f}^{(d)}$ based on a given value $\text{df}^{(d)}_{j,f}$. The orthogonalization involves the calculation of singular values $\bm{s}_j$ of a squared $M_j \times M_j$ matrix. Once $\bm{s}_j$ are computed, \eqref{eq:df} can also be solved multiple times for different df values without additional costs (see Supplement~\ref{app:dro} for details). Moreover, as the factorization only requires univariate smooth terms, $\lambda_{j,f}^{(d)}$ can be calculated for every feature separately at the cost of $\mathcal{O}(M_j^3)$ due to our factorization approach. This cost is comparatively small compared to a computation for all features in a $D$-variate interaction term with $\mathcal{O}(M^{3D})$. Also note that in \cref{th:homo}, $\text{df}_{j,f}^{(d)}(\lambda_{j,f}^{(d)})$ only needs to be calculated once for every $j$ as all involved matrices in \eqref{eq:df} are independent of $f$, and can be done prior to the optimization with no additional costs during training. 
\cref{alg:smoothing} summarizes the routine.
Homogeneous AHOFM smoothing amounts to equally flexible non-linear interactions for every order-$d$ AHOT and hence implies isotropic smoothing for all TPS. Given no a priori information on the non-linear interactions of all features, this is a natural choice. In contrast, if we choose different values for one or more features, i.e., $\exists j: \text{df}_{j,f} \neq \text{df}$, all TPS involving the $j$th feature are subject to anisotropic smoothing. 

\begin{figure}[!t]
    \begin{center}
\centerline{
    \includegraphics[width = \columnwidth]{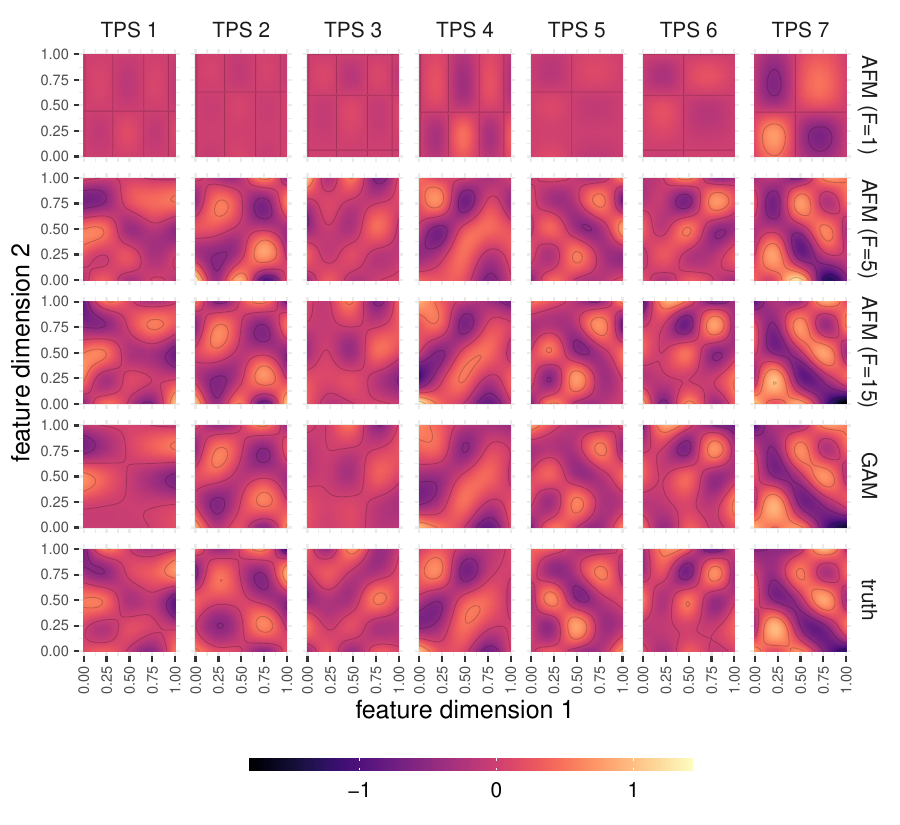}}
    \vskip -0.2in
    \caption{\small Estimated and true surfaces for 7 different TP splines (columns) and methods (rows) visualized by contour plots. Colors represent the partial effect values. }
    \label{fig:estsurf}
    \end{center}
    \vskip -0.3in
\end{figure}

\subsubsection{Optimization} \label{sec:optim}

In order to scale also for large numbers of observations, we propose a stochastic mini-batch gradient descent routine for the optimization of A(HO)FMs. We discuss the optimization problem of A(HO)FMs in Supplementary Material~\ref{app:opt} and suggest a block-coordinate descent as an alternative optimization routine by showing that the problem in \eqref{eq:penloss} is coordinate-wise convex in $\bm{\gamma}^{(d)}_{j,f}$ (Lemma~\ref{th:convex}). In practice, however, different block-coordinate descent variants showed slow convergence and finding a good choice for hyperparameters such as the learning rate proved to be challenging. In contrast, sophisticated stochastic gradient descent routines such as Adam \citep{Kingma.2014} showed similar or even better results. 
\section{NUMERICAL EXPERIMENTS}

We investigate the following research questions to empirically analyze theoretical findings and better understand the practical aspects of A(HO)FMs:
\vspace{-0.2cm}
\begin{enumerate}
    \item Do they provide a reasonable approximation of TPS surfaces (for increasing $F$)?
    \item Do they provide a reasonable prediction performance (for increasing $F$) within the class of GAMs?
    \item Can the theoretical model complexity be verified in practice?
    \item How well do they perform when applied to real-world data? Which are influencing factors for predictive accuracy and can we identify limitations when compared to alternative fitting methods?
\end{enumerate}
We will answer these questions in consecutive order in the following subsections numbered accordingly.


\begin{figure}[!t]
\begin{center}
\centerline{\includegraphics[width=0.85\columnwidth]{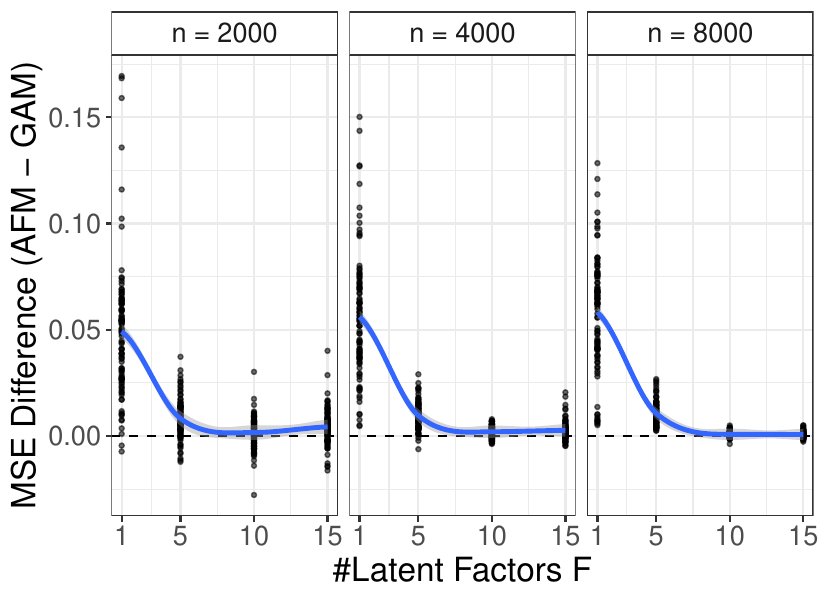}}
\vskip -0.15in
    \caption{\small Estimation quality measures by the MSE difference between a GAM estimation (gold standard) and our proposal with different numbers of latent dimensions $F$ (x-axis) and different numbers of observations (columns). Points correspond to different simulation replications and surfaces. A blue smoother function visualizes the trend in $F$. For smaller data sets, however, smaller $F$ can be beneficial by inducing additional regularization.}
    \label{fig:comp}
\end{center}
\vskip -0.3in
\end{figure}

\subsection{Estimation Performance} \label{sec:estperf}

We first compare the estimation performance of our proposal with the SotA for fitting GAMs with TPS. More specifically, we simulate features and generate bivariate non-linear effects for every possible feature pair. The response is generated by adding random noise with a signal-to-noise ratio of 0.5 to the sum of all bivariate effects. We then compare the estimation performance of all feature effects qualitatively by inspecting the estimated non-linear effects visually (cf. Figure~\ref{fig:estsurf} for $n=2000$), and quantitatively by computing the mean squared error (MSE) between the estimated and true surfaces (Figure~\ref{fig:comp}). For $n\in\{2000,4000,8000\}$ and $p=5$ (resulting in 10 bivariate effects), we run AFMs with $F\in\{1,5,15\}$. We repeat every setting 10 times with different random seeds. The GAM estimation can be thought of as a gold standard which is only subject to an estimation error, but no approximation error. In contrast, AFMs are also subject to an approximation error. Our quantitative analysis of results confirms this hypothesis. Figure~\ref{fig:comp} depicts the MSE differences for all analyzed settings to compare the estimation performance of AFMs and GAMs when calculating the average point-wise differences between the estimated bivariate surface and the true surface. Results suggest that for increasing $F$ our approach will approach the estimation performance of GAMs. With more data (larger $n$), this effect becomes even more clear.

\subsection{Prediction Performance} \label{sec:predperf}

\begin{figure}[!h]
\begin{center}
\centerline{\includegraphics[width=0.85\columnwidth]{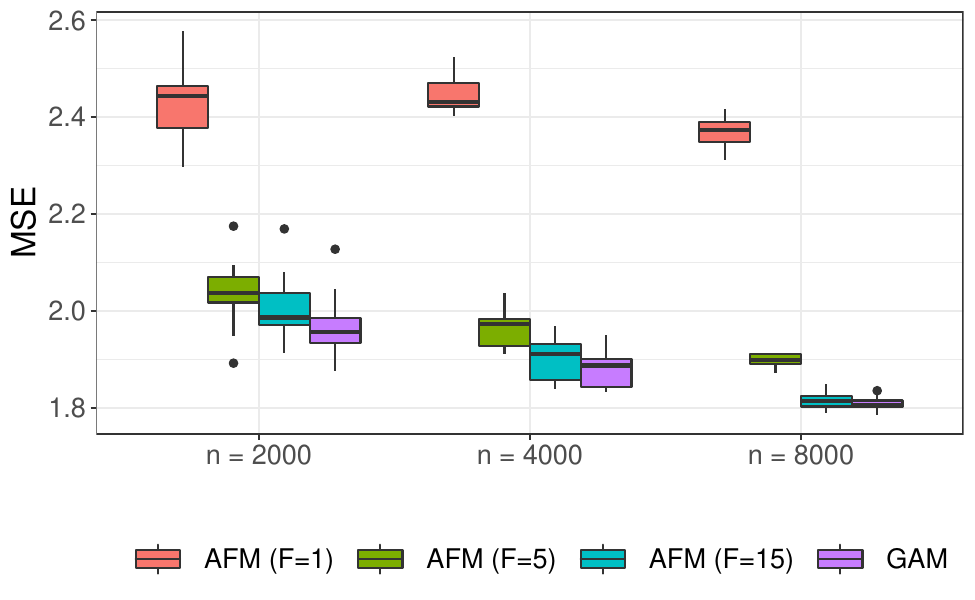}}
\vskip -0.15in
    \caption{\small Prediction error of the GAM (lower bound) and our proposal with different numbers of latent dimensions $F$ (colors) for different numbers of observations (x-axis).}
    \label{fig:compred}
\end{center}
\vskip -0.3in
\end{figure}

We follow the setup from the previous section and compare the prediction performance of the exact GAM and our approach to quantify the approximation error made by the factorization. Figure~\ref{fig:compred} depicts the results, confirming that the approximation error will tend to zero when increasing the number of observations $n$ or latent factors $F$.

\subsection{Scalability} \label{sec:memperf}

Our next experiment investigates the scaling behavior of our approach and compares it to the SotA implementation for big additive models \citep[BAMs][]{Wood.2017}. We simulate $p\in\{3,6,9,12\}$ standard normal distributed features for $n \in \{6000, 12000, 18000\}$ observations and fit both BAM and an AFM to learn a GAM with TP splines for all possible combinations of the $p$ features. Figure~\ref{fig:mem} (first page) summarizes the results by comparing the memory consumption (in megabytes) and the computing time (in seconds). The results reflect our initial motivation to propose AFMs. While both time and memory consumption grow exponentially in the number of features for BAMs, we observe a linear scaling for AFMs both for memory consumption and computation time.

\subsection{Benchmark Studies} \label{sec:bench}

\begin{table*}[!t]
\caption{Average test performance (MSE $\downarrow$) with standard deviation in brackets for FM methods (columns) and data sets (rows) over 10 different train-test splits. The best method per data set is highlighted in bold.}
\label{tab:benchmark1}
\begin{center}
\begin{footnotesize}
\begin{tabular}{lcccccc}
\toprule
 & GAM & FM & HOFM($D=3$) & AFM & AHOFM($D=3$) \\
\midrule
Airfoil  & 119.3 (2.490) &  90.70 (45.55)   & 79.03 (51.29)  & 
{4.314} (0.576) & \textbf{4.181} (0.699) \\
Concrete & 9.768 (8.408) & 10.55 (0.970)  & 10.56 (0.955)  & 
\textbf{6.100} (0.648)      &  {6.127} (0.491)   \\
Diabetes & 142.1 (9.626) & 143.7 (10.04) &  143.3 (9.312)  &  
\textbf{57.12} (6.243) & 63.65 (9.672) \\
Energy  & 3.409 (0.363) & 7.487 (0.620)   & 7.480 (0.632) &  
\textbf{3.137} (0.332)    & {3.183} (0.359)  \\
Forest Fire  &  1.465 (0.163)  &   {1.403} (0.105)  & \textbf{1.398} (0.107)  & 
1.573 (0.325)  & 1.773 (0.307) \\
Naval  & \textbf{0.002} (0.000) &  0.009 (0.001) & 0.009 (0.001) & 
0.004 (0.002) & {0.003} (0.001) \\
Yacht  &  3.027 (0.517) &   8.883 (0.990)    &    8.898 (0.985)  &  
{2.401} (0.618) & \textbf{1.743} (0.624) \\
\bottomrule
\end{tabular}
\end{footnotesize}
\end{center}
\vskip -0.1in
\end{table*}

To assess prediction performance and the {limitations} of our approach when applied to real-world data, we compare A(HO)FMs against a variety of alternative approaches with similar properties. In particular, all methods use linear or spline feature effects. The similarity of comparison methods allows us to objectively reason about our methods' pros and cons.  We compare all methods on commonly used benchmark data sets using 10 train-test splits and report average MSE values as well as their standard deviation. Further details on hyperparameters and benchmark data sets can be found in Section~\ref{app:bench} in the Supplementary Material.

\subsubsection{Ablation Study}

We first compare A(HO)FMs with their linear counterpart (HO)FMs and univariate GAMs. This study will provide statements to what extent A(HO)FMs' performance is driven by non-linearity and interactions. For this, we use three different numbers of latent dimensions ($F \in \{1,5,15\}$) and fix the degrees-of-freedom $df = 15$. We implement all methods in a unifying framework in TensorFlow \citep{Tensorflow}, use the Adam optimizer with default settings for optimization and find the best epoch via early stopping based on a 15\%-validation split. Table~\ref{tab:benchmark1} summarizes the results when choosing the best-performing hyperparameter set per method and data set. Our results show that A(HO)FMs perform well and indicate that the extension to non-linearities is in some cases beneficial (Energy, Naval, Yacht), and the combination of non-linearity and interaction can improve performance notably (for Airfoil, Concrete, Diabetes, and Yacht).

\subsubsection{Comparison with Sparse Approaches}

As A(HO)FMs model all splines and possible interactions thereof, the natural question arises if this can also be a limitation. To examine this question, we compare our approach against methods that still represent an additive smooth model but allow for variable selection. The results of this study thus additionally help in better understanding A(HO)FMs and can pave the way for future research by highlighting potential improvements to the presented approach. More specifically, we compare our method against boosting with univariate splines \citep[GAMBoost;][]{mboost} and boosting with univariate splines and bivariate tensor products (GA${}^2$MBoost). These methods optimize GAMs using a greedy coordinate-wise algorithm that inherently provides variable selection when used in combination with early stopping. We use the same settings as in the previous study for A(HO)FMs and select the best stopping iteration for boosting using also the same train/validation-split as for the FM approaches. Table~\ref{tab:benchmark2} summarizes the results.
\begin{table}[ht]
\caption{Average test performance (MSE $\downarrow$) with standard deviation in brackets for different methods (columns) and data sets (rows) over 10 different train-test splits. The best method is highlighted in bold.}
\label{tab:benchmark2}
\begin{center}
\begin{footnotesize}
\resizebox{1.0\columnwidth}{!}{
\begin{tabular}{lccc}
\toprule
  & GAMBoost & GA${}^2$MBoost  & A(HO)FM \\
\midrule
Airfoil  & 79.03 (51.29)  & 4.628 (0.213) & \textbf{4.181} (0.699) \\
Concrete    & 6.280 (0.486) & 6.292 (0.586)  & \textbf{6.100} (0.648)        \\
Diabetes   & \textbf{55.18} (5.020) & 55.16 (5.618) & 57.12 (6.243)  \\
Energy    & 3.780 (0.427) & 3.759 (0.368) & \textbf{3.137} (0.332)   \\
ForestF    & \textbf{1.388} (0.113) & 1.397 (0.098)  & 1.573 (0.325)  \\
Naval    & 0.012 (0.000) & \textbf{0.003} (0.001) &  \textbf{0.003} (0.001) \\
Yacht    & {1.532} (0.546) & \textbf{1.504} (0.504) & {1.743} (0.624) \\
\bottomrule
\end{tabular}
}
\end{footnotesize}
\end{center}
\vskip -0.15in
\end{table}
As before, A(HO)FMs perform well, in particular for data sets where interactions have been found beneficial in the previous ablation study (e.g., for Airfoil). Further, we see that boosting shows slightly better performance for three out of the 7 data sets, suggesting that feature selection can potentially improve prediction performance. As model-based boosting methods usually do not scale well with the number of features, a sparse A(HO)FM approach could thus be a promising future research direction. We note, however, that due to the nature of approximation used in A(HO)FMs (i.e., the shared latent factors across all interactions), incorporating a sparsity-inducing penalty is not straightforward.


\section{SUMMARY AND OUTLOOK}

We presented an additive model extension of HOFMs for scalable higher-order smooth function estimation based on TPS. The proposed approach allows fitting GAMs with $D$-variate smooth terms at costs similar to a univariate GAM. Our simulation studies showed that these machines approximate TPS surfaces very well and can match the exact GAM both in estimation and prediction performance. A(HO)FMs thereby not only allow to fit higher-order GAMs but also make additive models more competitive in their predictive performance shown in our benchmark study. 
A promising future research direction therefore is the sparsification of our proposed approach. This would not only allow for improving prediction performance as suggested by our benchmark study but is also closely related to the interpretability of A(HO)FMs, which we discuss in Supplement~\ref{app:interp} in more detail. A sparsification approach would require being compatible with the first-order optimization techniques used in this work. Hence, a possible interesting direction could be an optimization transfer \citep{chris}.


\bibliographystyle{plainnat}
\bibliography{sample-base}

\appendix

\onecolumn

\aistatstitle{Scalable Higher-Order Tensor Product Spline Models: \\
Supplementary Materials}
 
\section{Proofs} \label{app:proof}

\subsection{Proof of Lemma~\ref{th:afmrep}}
%
\begin{equation} \label{eq:afmrepproof}
\begin{split}
&\quad \sum_{k=1}^p \sum_{l=k+1}^{p} \sum_{m=1}^{M_k} \sum_{o=1}^{O_l} B_{m,k}(x_k) B_{o,l}(x_l) \beta_{m,k,o,l}\\
&\approx 
\sum_{k=1}^p \sum_{l=k+1}^{p} \sum_{m=1}^{M_k} \sum_{o=1}^{O_l} \sum_{f=1}^F B_{m,k}(x_k) B_{o,l}(x_l) \gamma_{m,k,f} \gamma_{o,l,f}\\
& = \sum_{k=1}^p \sum_{l=k+1}^{p} \sum_{m=1}^{M_k} \sum_{o=1}^{O_l} \sum_{f=1}^F B_{m,k}(x_k) \gamma_{m,k,f} B_{o,l}(x_l)  \gamma_{o,l,f}\\
& =  \sum_{f=1}^F \left\lbrace \sum_{k=1}^p  \sum_{m=1}^{M_k}  B_{m,k}(x_k) \gamma_{m,k,f} \sum_{l=k+1}^{p} \sum_{o=1}^{O_l} B_{o,l}(x_l)   \gamma_{o,l,f} \right\rbrace\\
& =  \sum_{f=1}^F \left\lbrace \sum_{k=1}^p  \sum_{m=1}^{M_k}  B_{m,k}(x_k) \gamma_{m,k,f} \sum_{l=k+1}^{p} \varphi_{l,f} \right\rbrace\\
& =  \sum_{f=1}^F \left\lbrace \sum_{k=1}^p  \sum_{m=1}^{M_k}  \sum_{l=k+1}^{p}  \underbrace{B_{m,k}(x_k) \gamma_{m,k,f}}_{c_{m,k,f}} \varphi_{l,f} \right\rbrace\\
%
& = \frac{1}{2} \sum_{f=1}^F \left\lbrace \sum_{k=1}^p \sum_{m=1}^{M_k} \sum_{l=1}^{p} c_{m,k,f} \varphi_{l,f} - \sum_{k=1}^p   \sum_{m=1}^{M_k}   c_{m,k,f}  \varphi_{k,f} \right\rbrace \\
& = \frac{1}{2} \sum_{f=1}^F \left\lbrace  \left\lbrack \sum_{k=1}^p \sum_{m=1}^{M_k} c_{m,k,f} \right\rbrack  \left\lbrack \sum_{l=1}^{p} \sum_{o=1}^{O_l}  c_{o,l,f} \right \rbrack - \sum_{k=1}^p \varphi_{k,f}^2 \right\rbrace \\
& = \frac{1}{2} \sum_{f=1}^F \left\lbrace  \left\lbrack \sum_{k=1}^p \sum_{m=1}^{M_k} c_{m,k,f} \right\rbrack^2  - \sum_{k=1}^p  \varphi_{k,f}^2 \right\rbrace \\
&= \frac{1}{2} \sum_{f=1}^F \left\lbrace  \left\lbrack \sum_{k=1}^p \varphi_{k,f} \right\rbrack^2  - \sum_{k=1}^p \varphi_{k,f}^2 \right\rbrace.
\end{split}
\end{equation}

\subsection{Proof of \cref{th:afmscale} and~\ref{th:afmstore}}

Both propositions directly follow from the fact that \eqref{eq:afmrepproof} only sums over $f$, $k$ and $m$ once, and every basis function $B_{m,k}$ is therefore also only evaluated once.

\subsection{Proof of \cref{th:ahot}} \label{app:proofahot}

An alternative representation of \eqref{th:ahot} is given by
\begin{equation} \label{eq:AHOT2}
\Phi^{(d)}_f = \sum_{j_d > \cdots > j_1} \prod_{t=1}^d \varphi_{j_t,f}
\end{equation}
by just plugging in the definition for $\varphi$. We can consider \eqref{eq:AHOT2} as an ANOVA kernel of degree $d$ in the new feature space given by all $\varphi$s. As a result, the multi-linearity property of the ANOVA kernel holds \citep[see][Appendix B.1]{blondel2016higher}, i.e.,
\begin{equation} \label{eq:multilin}
    \Phi^{(d)}_f = \Phi^{(d)}_{f,\neg j} + \varphi_{j,f} \Phi^{(d-1)}_{f,\neg j}
\end{equation}
where $\Phi^{(d)}_{f,\neg j} = \sum_{\{j_d > \cdots > j_1\} \backslash j} \prod_{t=1}^d \varphi_{j_t,f}$ and therefore AHOFMs can be represented as in \cref{th:ahot} by using the same arguments as for HOFMs \citep{blondel2016higher}.

\subsection{Proof of \cref{th:convex}}

For simplicity assume that the model predictor only consists of a single AHOT, i.e., $\eta(\bm{x}) = \sum_{f=1}^{F_d} \Phi_f^{(d)}$. The generalization of the following statement to several AHOTs follows due to the additivity of the model predictor. 
Using \cref{eq:multilin} and constants $\xi_f, \zeta_f$, it follows
\begin{equation} \label{eq:affinity}
\begin{split}
    &\eta(\bm{x}) = \sum_{f=1}^{F_d} \Phi_f^{(d)} 
    = \sum_{f=1}^{F_d} \left\{  \Phi^{(d)}_{f,\neg j} + \varphi_{j,f} \Phi^{(d-1)}_{f,\neg j} \right\}
    = \sum_{f=1}^{F_d} \left\{ \xi_f + \varphi_{j,f} \zeta_f \right\} \\
    &= \sum_{f=1}^{F_d} \left\{ \xi_f + \bm{B}_{j}^\top\bm{\gamma}_{j,f} \zeta_f  \right\} = \sum_{f=1}^{F_d} \xi_f + (\bm{B}_{j}\otimes \bm{\zeta})^\top\bm{\Gamma}^{(d)}_{j} = \text{const.} + \langle \tilde{\bm{B}}_{j}, \bm{\Gamma}^{(d)}_{j} \rangle, 
\end{split}
\end{equation}
where ${\bm{\Gamma}}^{(d)}_{j} = \text{vec}(\mathfrak{G}^{(d)}_{:,j,:}) \in \mathbb{R}^{MF_d}$, $\tilde{\bm{B}} = \bm{B}_{j}\otimes \bm{\zeta} \in \mathbb{R}^{MF_d}$ and $\bm{\zeta} = (\zeta_1,\ldots,\zeta_{F_d}) \in \mathbb{R}^{F_d}$. \eqref{eq:affinity} shows that $\eta$ is an affine function in $\bm{\Gamma}^{(d)}_{j}\,\forall j\in[p]$.
Now let $\ell$ be a convex loss function of $\eta$ and note that $\mathcal{P}(\mathfrak{G}, \bm{\Theta})$ in \cref{def:ahofmpen} is decomposable across the parameters $\bm{\gamma}^{(d)}_{j,f}$. The composition of $\ell$ and $\eta$ is convex and as $\mathcal{P}$ is decomposable in $\bm{\gamma}^{(d)}_{j,f}$ and hence also in $\bm{\Gamma}^{(d)}_j$, the penalized objective in \eqref{eq:penloss} is convex w.r.t.~every $\bm{\Gamma}^{(d)}_j$ and thus every $\bm{\gamma}_{j,f} \, \forall j \in [p], f \in [F_d]$.

\section{Optimization} \label{app:opt}

Having defined the objective in \eqref{eq:penloss}, we obtain the following result. 
\begin{lemma} \label{th:convex}
The optimization problem in \eqref{eq:penloss} is coordinate-wise convex in $\bm{\gamma}^{(d)}_{j,f}$.
\end{lemma}
A corresponding proof is given in Appendix~\ref{app:proof}. Using this finding suggests a block coordinate descent (BCD) solver as an alternative approach to optimize (penalized) AHOFMs with block updates for $\bm{\gamma}^{(d)}_{j,f}$. In contrast to (HO)FMs, we perform block updates instead of plain coordinate descent as the AHOFM penalty is only decomposable w.r.t.~all $\bm{\gamma}^{(d)}_{j,f}$, but not w.r.t.~$\gamma^{(d)}_{m,j,f}$. For BCD we require several quantities:
\begin{equation} \label{eq:quants}
\begin{split}
\nabla \varphi_{j,f}(x_{i,j}) &:= \partial \varphi_{j,f}(x_{i,j}) / \partial \bm{\gamma}_{j,f} = \bm{B}_j(x_{i,j}),\\
  \nabla \Phi_f^{(d)}(\bm{x}_i) &:= \frac{\partial \Phi_f^{(d)}(\bm{x}_i)}{\partial \bm{\gamma}^{(d)}_{j,f}}\\
  &= \frac{1}{d} \sum_{t=1}^d (-1)^{t+1} \left\{ \frac{\partial \Phi_f^{(d-t)}(\bm{x}_i)}{\partial \bm{\gamma}^{(d)}_{j,f}} \left\{ \sum_{j=1}^p \left[\varphi^{(d)}_{j,f}(x_{i,j})\right]^t \right\} \right.\\
  &+ \left. \Phi_f^{(d-t)(\bm{x}_i)} t \left[\varphi^{(d)}_{j,f}(x_{i,j})\right]^{t-1} \nabla \varphi_{j,f}(x_{i,j}) \right\},\\
  \nu &= \left\{  \sum_{i=1}^n \frac{\partial^2 \ell(\bm{x}_i,y_i)}{\partial (\bm{\gamma}^{(d)}_{j,f})^2} + \lambda^{(d)}_{j,f} \bm{P}_j  \right\}^{-1}\\
  \frac{\partial \mathcal{L}(\bm{x}_i,y_i)}{\partial \bm{\gamma}^{(d)}_{j,f}} &= \frac{\partial \ell(\bm{x}_i,y_i)}{\partial \bm{\gamma}^{(d)}_{j,f}} \nabla \Phi_f^{(d)}(\bm{x}_i)  +  \lambda^{(d)}_{j,f} \bm{P}_j \bm{\gamma}^{(d)}_{j,f}.
\end{split}
\end{equation}
Note that the first term is involved in every update step, but independent of $f$ and the iteration. It is therefore possible to cache the result once at the beginning of the training routine as also mentioned in \cref{alg:init}. For reverse-mode differentiation, note that the second term can be calculated efficiently by caching intermediate results.

A high-level routine is described in \cref{alg:ahofm} (for simplicity for the case with only a single $D$-variate smooth). 
In \cref{alg:ahofm}, we require gradients
\begin{equation*}
\nabla \mathcal{L}(\bm{\gamma}^{(d)}_{j,f}) := \sum_{i=1}^n \frac{\partial \mathcal{L}(\bm{x}_i, y_i)}{\partial \bm{\gamma}^{(d)}_{j,f}},
\end{equation*}
where $\mathcal{L}$ is the objective function from \eqref{eq:penloss} and a learning rate $\nu$, which is defined above. Various terms involved in the update step can be pre-computed or cached (see \cref{app:alg} for details).
\begin{algorithm}[htbp]
   \caption{BCD AHOFM Optimization}
   \label{alg:ahofm}
\begin{algorithmic}
   \STATE {\bfseries Input:} Data $(y_i,\bm{x}_i)_{i\in[n]}$; $B_{m,j},\bm{P}_j, \,\forall j\in[p],m\in[M_j]$; $D$; $\text{df}^{(D)}$; $F_D$; BCD convergence criterion
   \STATE {\bfseries Initialization:} $\hat{\eta}, \mathfrak{G}^{(D)} = \text{init}(Input)$ (Appendix~\ref{app:init})
   \REPEAT
   \FOR{$f=1$ {\bfseries to} $F_D$}
   \FOR{$j=1$ {\bfseries to} $p$}
   \STATE Calculate step-size $\nu$ 
   \STATE Update $\bm{\gamma}^{(d)}_{j,f} \leftarrow \bm{\gamma}^{(d)}_{j,f} - \nu \nabla \mathcal{L}(\bm{\gamma}^{(d)}_{j,f})$
   \STATE Synchronize ${\hat{\eta}}_i, i\in[n]$
   \ENDFOR
   \ENDFOR
   \UNTIL{convergence}
\end{algorithmic}
\end{algorithm}

\section{Algorithmic Details} \label{app:alg}

\subsection{Demmler-Reinsch Orthogonalization} \label{app:dro}

We here describe the Demmler-Reinsch Orthogonalization (DRO; \cref{alg:dro}) and sv2la (\cref{alg:sv2la}) routine proposed to efficiently compute smoothing penalties. Details can be found in \cite{ruppert2003semiparametric}, Appendix B.1.1. We use $\text{Chol}$ to denote the Cholesky decomposition of a matrix and $\text{SVD}$ for the singular value decomposition of a matrix.

\begin{algorithm}[htbp]
   \caption{DRO}
   \label{alg:dro}
\begin{algorithmic}
   \STATE {\bfseries Input:} Feature matrix $\bm{B} \in \mathbb{R}^{n\times M}$, penalty matrix $\bm{P}\in\mathbb{R}^{M\times M}$
   \STATE Compute:
   \begin{enumerate}
       \item $\bm{R}^\top \bm{R} \leftarrow \text{Chol}(\bm{B}^\top\bm{B})$
       \item $\bm{U}\text{diag}(\bm{s})\bm{U}^\top  \leftarrow \text{SVD}(\bm{R}^{-T}\bm{P}\bm{R}^{-1})$
   \end{enumerate}
    \STATE {\bfseries Output:} singular values $\bm{s}$
\end{algorithmic}
\end{algorithm}

\begin{algorithm}[htbp]
   \caption{sv2la}
   \label{alg:sv2la}
\begin{algorithmic}
   \STATE {\bfseries Input:} Singular values $\bm{s} \in \mathbb{R}^M$, $\text{df}$
   \STATE Define $\text{dffun}(l) = \sum_{j=1}^M (1+ls_j)^{-1}$;
   \STATE Compute: $\lambda$ for which $\text{dffun}(\lambda) = \text{df}$ using a uniroot search;
      \STATE {\bfseries Output:} $\lambda$
\end{algorithmic}
\end{algorithm}

\subsection{{Homogeneous AHOFM Smoothing}}

Given the previous algorithms, we can implement homogeneous AHOFM smoothing as described in Algorithm~\ref{alg:smoothing}.

\begin{algorithm}[htbp]
   \caption{Homogeneous AHOFM Smoothing}
   \label{alg:smoothing}
\begin{algorithmic}
   \STATE {\bfseries Input:} $\bm{B}_j, \bm{P}_j\,\forall j\in[p]$; $\text{df}^{(d)} \,\forall d\in[D]$
   \FOR{$j=1$ {\bfseries to} $p$}
   \STATE Compute $\bm{s}_j = \text{DRO}(\bm{B}_j,\bm{P}_j)$ (costs: $\mathcal{O}(M_j^3)$)
   \FOR{$d=1$ {\bfseries to} $D$}
   \STATE Compute $\lambda_{j,1}^{(d)} = \text{sv2la}(\bm{s}_j,\text{df}^{(d)})$ (negligible costs);
   \STATE Set $\lambda_{j,f}^{(d)} = \lambda_{j,1}^{(d)}$ for $f\in[F_d]$;
   \ENDFOR
   \ENDFOR
      \STATE {\bfseries Output:} $\lambda_{j,f}^{(d)}$ for all $j\in[p],d\in[D],f\in[F_d]$
\end{algorithmic}
\end{algorithm}

\subsection{AHOFM Initialization} \label{app:init}

Putting everything together, the initialization of AHOFMs is given in Algorithm~\ref{alg:init}.

\begin{algorithm}[htbp]
   \caption{AHOFM Init}
   \label{alg:init}
\begin{algorithmic}
\STATE {\bfseries Input:} Data $(y_i,\bm{x}_i), i\in[n]$; order $D$, bases functions $B_{m,j},\bm{P}_j, j\in[p],m\in[M_j]$; $\text{df}^{(d)}$
   \STATE Compute the following quantities:
   \begin{itemize}
       \item $\lambda_{j,f}^{(d)} \,\forall j\in[p],f\in[F_d]$ using \cref{alg:smoothing};
       \item $\nabla \varphi_{j,f}(x_{i,j}) \,\forall i\in[n]$ as in \eqref{eq:quants};
   \end{itemize}
   Cache the derivatives $\nabla \varphi_{j,f}(x_{i,j})$ for later update steps;
   \STATE Randomly initialize $\bm{\gamma}^{(d)}_{j,f}$ and calculate $\varphi^{(d)}_{j,f}$ for all $j\in[p],f\in[F_d]$;
   \STATE Compute $\hat{\eta}$
   \STATE {\bfseries Output:} $\hat{\eta}, \mathfrak{G}^{(d)}$ 
\end{algorithmic}
\end{algorithm}

\section{Interpretability of A(HO)FMs} \label{app:interp}

\begin{figure}[ht]
    \begin{center}
\centerline{
    \includegraphics[width = 0.5\columnwidth]{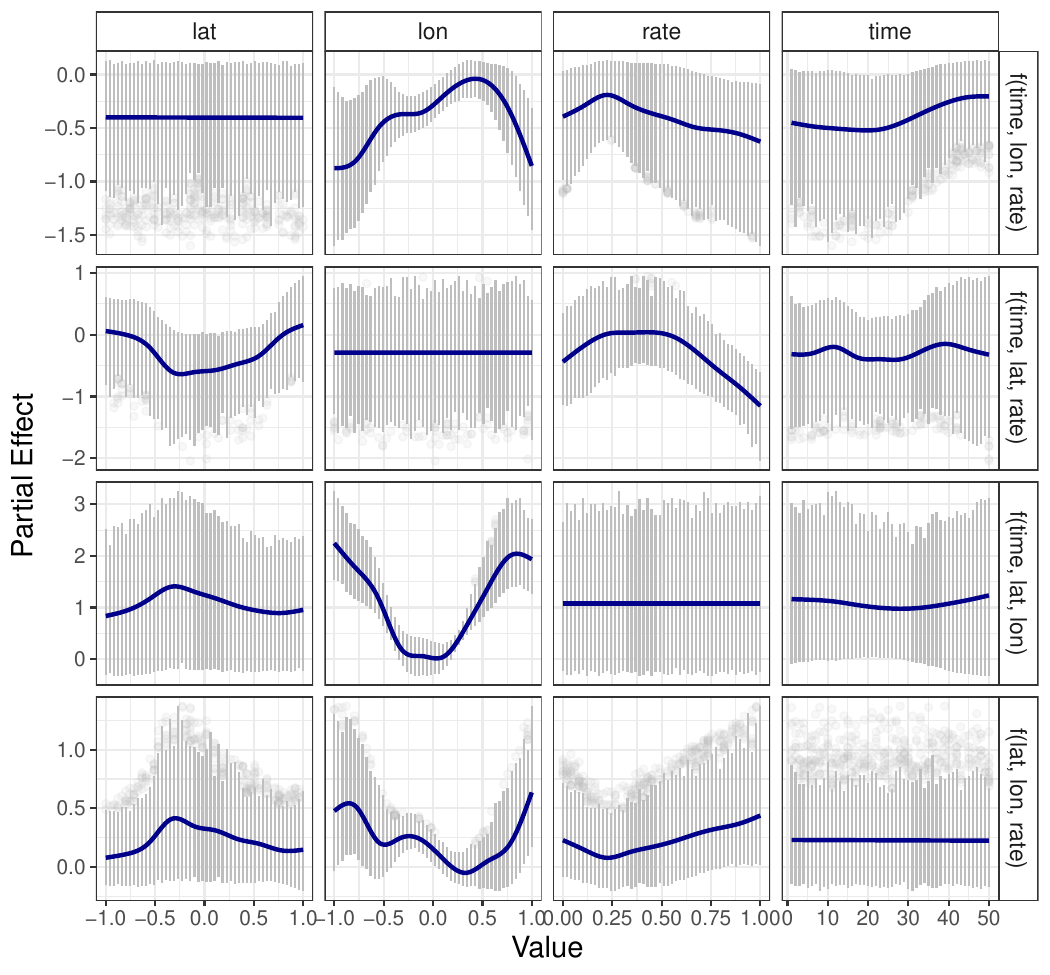}}
    \vskip -0.1in
    \caption{Estimated partial effects of different features (columns) for different three-dimensional functions (rows). Blue lines indicate the marginal average in the respective feature direction while gray vertical lines show the spread across the other two dimensions. More variation indicates larger variation across the other two dimensions. Features not involved in a partial effect (diagonal from top left to bottom right) naturally have a constant effect.}
    \label{fig:interp}
    \end{center}
\end{figure}

Due to their additivity assumption, every additive feature effect in GAMs can be interpreted on its own (ceteris paribus). Although AHOFMs inherit some of the interpretability properties from GAMs, e.g., their additivity, interpreting higher-order (non-linear) interaction terms remains challenging and cannot be done without considering lower-order effects of the same feature. For larger values of $p$, the quickly growing number of additive terms further makes it infeasible to grasp the influence of certain features or interactions. While this is a limitation of the current approach, we here propose three ways to check effects for models with small to moderate $p$. The first approach examines interaction terms by visualizing the single univariate smooth terms $\varphi^{d}_{j,f}$ for $f=1,\ldots,F^d$ and all involved feature dimensions $j$. Analyzing the univariate latent dimensions separately is not a new approach and the use of factorization approaches can even be motivated by the need to interpret higher-dimensional interactions in lower dimensions \citep[see, e.g.,][]{stocker2021functional}. 
Another approach that focuses only on the interaction effects itself is to visualize the actual approximations $\phi_{j_1,\ldots,j_d}$ as defined in \eqref{eq:afmapprox}. This reduces the number of terms to analyze by the factor $F_d$, but requires a method for presenting the $d$-variate effect. As shown in Figure~\ref{fig:interp}, a third approach is to visualize the marginals of these multivariate functions together with their variation across the respective other dimensions.

\paragraph{Experiments} We simulate a toy example for $D=3$ with four features to demonstrate the third approach. For better understanding, the features are referred to as \emph{time}, \emph{lat}, \emph{lon} and \emph{rate}. The outcome is assumed to be normally distributed with $\sigma = 0.1$ and the mean given by the sum of all possible smooth 3-way interactions of the features. To simulate non-linear three-dimensional functions, we use a basis evaluation of features with 4 degrees-of-freedom for \emph{time}, 5 degrees-of-freedom for \emph{lat}, 7 degrees-of-freedom for \emph{lon} and 5 degrees-of-freedom for \emph{rate}. Partial effects for each three-dimensional smooth are generated by calculating the TP for these basis and randomly drawing coefficients for the resulting TPS. We generate $10^4$ observations, fit a AHOFM($D=3$) and visualize the resulting effects in two different ways. 
Figure~\ref{fig:interp} depicts the marginal univariate effects of all four features for all four 3-way interaction effects. This allows us to see how each feature marginally affects each of the interaction terms. Additionally, it shows how much variation the marginal effects have in the respective other two dimensions and thereby provides information on how much the features interact with the respective two other variables. 


\section{Numerical Experiments Details} \label{app:exp}

\subsection{Implementation}

All methods have been implemented in TensorFlow \citep{Tensorflow} using the package deepregression \citep{deepregression}, except for GAMBoost, where we used the package mboost \citep{mboost} and the SotA GAM, implemented in mgcv \citep{Wood.2017.book}.





\subsection{Benchmark Details} \label{app:bench}


As described in Section~\ref{sec:bench}, neither GAMs, (HO)FMs nor AHOFMs have many hyperparameters to tune. We investigate the influence of the latent dimension by testing $F=1,5,10$ for all approaches and, for a fair comparison between GAMs and A(HO)FMs, set the $df$ values for all methods to the same value $15$. 
All methods use early stopping on 15\% validation data with a patience of 50.

Table~\ref{tab:further} further lists the data characteristics for our benchmark data. Pre-processing is only done for ForestF, using a logp1 transformation for \texttt{area} and a numerical representation for \texttt{month} and \texttt{day}. 

\begin{table}[]
\begin{footnotesize}
\begin{center}
\caption{Data set characteristics, additional pre-processing and references.} \label{tab:further}
\begin{tabular}{cccc}
Data set & \# Obs. & \# Feat. & Reference \\ \hline 
Airfoil & 1503 & 5 &  \cite{Dua.2019}  \\
Concrete   & 1030 & 8      &    \cite{Yeh.1998} \\
Diabetes  & 442 & 10      &       \cite{Efron.2004} \\
Energy  & 768 & 8           &  \cite{Tsanas.2012} \\
ForestF   & 517 & 12 &    \cite{Cortez.2007} \\
Naval & 11934 & 16 &  \cite{Coraddu.2013}\\
Yacht  & 308 & 6           &    \cite{Ortigosa.2007,Dua.2019} \\ \hline
\end{tabular}
\end{center}
\end{footnotesize}
\end{table}

\end{document}